\documentclass[runningheads]{llncs}

\usepackage{eccv}

\usepackage[table]{xcolor}

\usepackage{eccvabbrv}
\usepackage[export]{adjustbox} 
\usepackage{fontawesome5} 
\usepackage{tcolorbox}
\tcbuselibrary{skins}
\usepackage{enumitem}
\usepackage{xcolor}
\tcbuselibrary{breakable} 
\definecolor{promptheader}{RGB}{50,50,50} 
\definecolor{promptbody}{gray}{0.95}      
\usepackage{graphicx}
\usepackage{booktabs}

\usepackage[accsupp]{axessibility}  

\usepackage{hyperref}

\usepackage{orcidlink}

\begin{document}

\title{ProMMSearchAgent: A Generalizable Multimodal Search Agent Trained with Process-Oriented Rewards}

\author{Wentao Yan\inst{1} \and
Shengqin Wang\inst{1} \and
Huichi Zhou\inst{3,5} \and
Yihang Chen\inst{3,5} \and
Kun Shao\inst{4} \and
Yuan Xie\inst{1,2} \and
Zhizhong Zhang\inst{1}\thanks{Corresponding author.}}

\institute{East China Normal University \and
Shanghai Innovation Institute \and
Huawei Noah's Ark Lab \and
Independent Researcher \and
University College London \\
\email{zzzhang@cs.ecnu.edu.cn}}
\titlerunning{ProMMSearchAgent}
\authorrunning{W.~Yan et al.}

\maketitle

\begin{abstract}
Training multimodal agents via reinforcement learning for knowledge-intensive visual reasoning is fundamentally hindered by the extreme sparsity of outcome-based supervision and the unpredictability of live web environments. To resolve these algorithmic and environmental bottlenecks, we introduce ProMMSearchAgent, establishing a novel Sim-to-Real training paradigm for multimodal search.

We decouple policy learning into a deterministic, local static sandbox. Crucially, to learn effectively within this constrained environment, we propose an introspective process-oriented reward. By probing the agent's own parametric knowledge boundaries, we generate dense behavioral metadata that explicitly rewards the correct cognitive decision, initiating a multimodal or text search only when visually or factually uncertain. Extensive experiments demonstrate that our locally-trained policy transfers zero-shot to the live Google Search API. ProMMSearchAgent achieves new SOTA performance, outperforming MMSearch-R1 by +5.1\% on FVQA-test, +6.3\% on InfoSeek, and +11.3\% on MMSearch.
\end{abstract}

\section{Introduction}
\label{sec:intro}
\begin{figure}[t]
    \centering
    \includegraphics[width=0.7\textwidth]{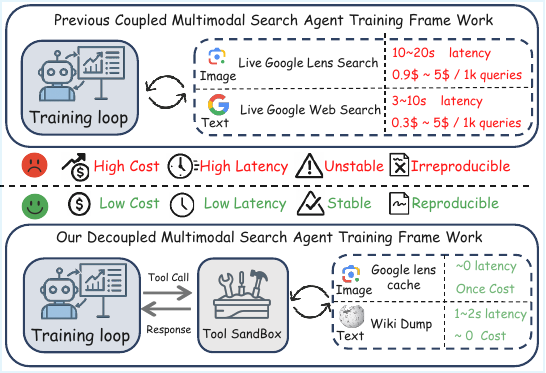} 
    \caption{Training Framework Comparison. (Top) The previous coupled framework relies on live APIs, resulting in High Cost, High Latency, Instability, and Irreproducibility. (Bottom) Our decoupled framework uses a Tool Sandbox with a local cache and Wiki dump, enabling Low Cost, Low Latency, Stable, and Reproducible training.}
    \label{fig:1}
\end{figure}

Multimodal large language models (MLLMs) \cite{openai2024gpt4ocard,liu2024improved,chen2024internvl,liu2023visual,li2024llava,li2024aria} have made rapid progress in vision-language reasoning. However, as these models transition from static parametric knowledge bases into dynamic, agentic systems \cite{wu2025mmsearch,geng2025webwatcherbreakingnewfrontier}, they encounter severe bottlenecks in multimodal deep search tasks. Solving knowledge-intensive visual question answering (VQA)—such as identifying an uncommon landmark or verifying time-sensitive visual facts—requires agents to iteratively formulate queries, often utilizing multimodal tools like image-to-image search to ground specific visual entities, translating these visual groundings into targeted text queries, and synthesizing the retrieved evidence.

Optimizing MLLMs for such complex, multi-step search behaviors is hindered by the extreme sparsity of effective supervision signals. Since standard reinforcement learning (RL) approaches rely predominantly on final outcome rewards, models often develop a degenerate strategy: indiscriminately invoking search tools to maximize the chance of stumbling upon the right answer. To prevent models from over-relying on external tools, existing methods like MMSearch-R1 \cite{wu2025mmsearch} introduce a heuristic "tool penalty" to discourage tool calls. However, our empirical analysis reveals that this introduces a critical \emph{reward-action mismatch}: for visually ambiguous samples that inherently require external search, the agent is penalized for taking the correct action. This paradox entangles answer correctness with proper tool-use strategy, leaving the agent unable to reliably learn its own visual knowledge boundaries.

To resolve this paradox and enable the agent to reliably identify its knowledge boundaries, we propose an introspective Process-Oriented Reward for multimodal reasoning. We define the ideal agentic behavior as a hierarchical "self-query" within the Chain-of-Thought (CoT). Specifically, the model must evaluate its self confidence in both visual grounding and factual knowledge before acting. If a knowledge gap is detected in either modality, it must trigger the appropriate tool; otherwise, it provides a direct answer. To train this tool-use decision-making, we introduce an offline probing step to map the model's true parametric limits. Specifically, prior to RL training, we conduct multiple tool-free inference rollouts for each training sample. If the model consistently fails to answer a query across these independent rollouts, it provides empirical proof of a genuine knowledge gap, and we annotate the sample with a 'search-required' behavioral tag. During RL, the agent receives an explicit positive reward if its chosen macro-action (tool vs. tool-free) aligns with this empirically derived metadata. By reinforcing this consistency, we ensure the model learns an invariant search logic calibrated to its own internal knowledge gaps.

Crucially, optimizing this fine-grained RL policy requires extensive exploration, which exposes a severe systemic bottleneck. As illustrated in Figure \ref{fig:1}, current training paradigms rely heavily on live web APIs, making the training process computationally high-cost, high-latency, unstable, and notoriously irreproducible. Training an RL agent against such a moving target makes rigorous ablation studies and experimental reproducibility virtually impossible. 

To provide a deterministic optimization landscape, we propose ProMMSearchAgent, introducing a Sim-to-Real decoupled training paradigm. We construct a functionally equivalent, controlled simulation environment (a Tool Sandbox) utilizing a local static corpus and a pre-computed multimodal cache. Fundamentally, the success of this Sim-to-Real transfer hinges on the synergy between the deterministic sandbox and our process-oriented reward. In a constrained local corpus, many complex queries are inherently unsolvable. However, our Process-Oriented Reward transforms the sandbox from a restricted cage into a focused "gymnasium": by rewarding the logical behavior of query formulation rather than the retrieval outcome, the agent learns a robust, invariant search policy. The sandbox provides the stable environment needed for precise credit assignment, while the process reward ensures the agent values its search actions as high-utility cognitive tools, regardless of whether the local corpus contains the final answer.

By rectifying the optimization landscape through this process-reward and sandbox synergy, ProMMSearchAgent learns a robust, invariant search strategy. We demonstrate that the policy learned in this localized, static environment transfers zero-shot to the live Google Search API at test time. Extensive experiments on knowledge-intensive VQA benchmarks reveal that ProMMSearchAgent not only achieves SOTA accuracy but also exhibits profound adaptive decision-making. Specifically, the agent tends to invoke precise multimodal or unimodal tools when it identifies a knowledge gap, while largely abstaining from external queries when its internal parametric knowledge is sufficient. This improved calibration suggests that our decoupled training paradigm effectively fosters a degree of autonomous search intelligence.

Our contributions are summarized as follows:
\begin{itemize}
\item We identify the \emph{reward-action mismatch} and supervision sparsity in current MLLM agents, revealing how flawed search penalties inhibit the learning of proper knowledge boundaries.
\item We propose an introspective \emph{process-oriented reward} that elevates self-probed behavioral metadata into an explicit dense RL signal, effectively resolving the credit assignment problem in search trajectories.
\item We construct a novel, difficulty-graded multimodal VQA dataset designed explicitly as a robust training testbed. This dataset is specifically curated to probe knowledge-dependent search decisions, providing the essential curriculum needed to cultivate the agent's hierarchical cognitive reasoning.
\item We introduce a \emph{Sim-to-Real} decoupled training paradigm using a static sandbox. We demonstrate that the learned invariant search logic generalizes zero-shot to live web APIs, achieving SOTA performance while enabling low-cost, fully reproducible agent training.
\end{itemize}

\section{Related Works}
\textbf{Reinforcement Learning.} 
Reinforcement Learning (RL) offers a novel paradigm for enhancing the reasoning capabilities of language models. In 2024, OpenAI's o1 and DeepSeek-R1 achieved significant performance breakthroughs in text reasoning tasks through a hybrid training strategy combining “Chain of Thought”\cite{hu2023avis,zhang2024visionsearchassistantempower} and RL\cite{jaech2024openai,guo2025deepseek,chen2025learning}. This achievement has inspired advancements in multimodal modeling. For instance, Apple Machine Learning Team proposed embedding MLLMs into embodied systems (e.g., robots, interactive interfaces) to enhance environmental perception and action capabilities through action expert distillation\cite{peng2024grounding}. Existing RL methods primarily rely on outcome-based supervision, where reward signals are constructed based on the correctness of final answers. This approach struggles to directly evaluate the quality of intermediate reasoning steps, potentially leading to biased optimization directions\cite{deng2025openvlthinker}. 

\textbf{Search-oriented Reasoning.}
Search-oriented reasoning has emerged as a key direction in multimodal modeling. Traditional retrieval-augmented generation (RAG) relies on static knowledge bases\cite{jiang2025rationalitylanguagemultimodalagents,zheng2025deepresearcher}. For instance, Search-R1 first employs reinforcement learning to train models to autonomously generate search queries, enabling multi-turn interactive reasoning and search\cite{jin2025search,li2025webthinkerempoweringlargereasoning,wu2025webwalkerbenchmarkingllmsweb}. In multimodal settings, MMSearch-R1 extends this idea by training an MLLM to invoke online search tools on demand, significantly improving visual reasoning on knowledge-intensive VQA benchmarks. DeepMMSearch-R1\cite{narayan2025deepmmsearch} further extends this direction by incorporating a croppable image search tool based on Grounding DINO\cite{liu2024groundingdinomarryingdino}. It is optimized with a two-stage pipeline that combines supervised finetuning on a dedicated DeepMMSearchVQA dataset with online RL over live web search tools. WebWatcher targets open-web “deep research” scenarios, using synthetic browsing trajectories and RL to train a multimodal agent that navigates real web pages and tools, such as web visit and Python interpreter. These agents demonstrate the importance of explicitly training models to use multimodal search in a multi-step fashion.

\section{Method}
\label{sec:method}

\subsection{Behavioral Metadata Engine}
\label{sec:data}
\begin{figure}[t]
    \centering
    \includegraphics[width=0.75\textwidth]{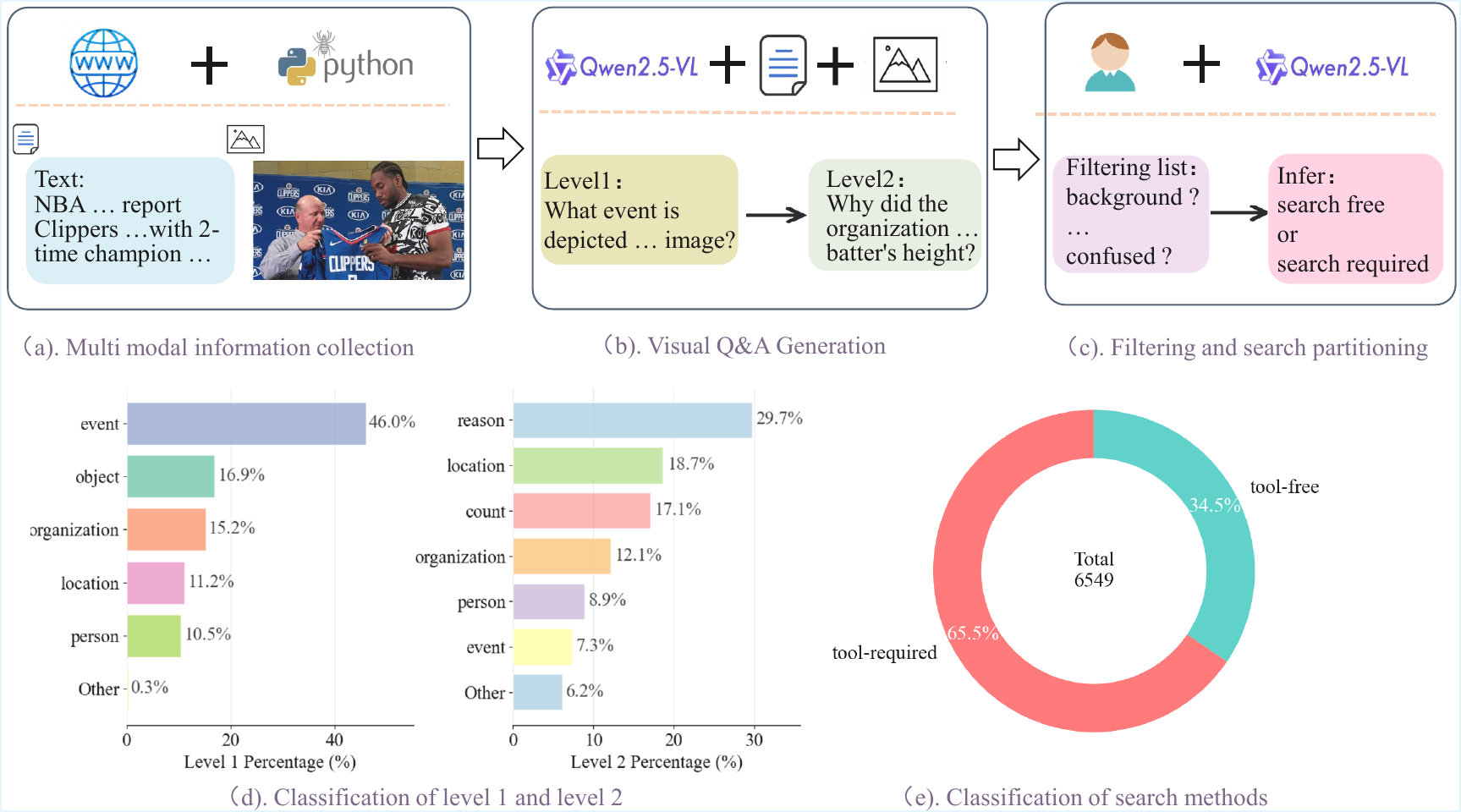} 
    \caption{Illustration of data construction process of our datasets: (a). Website information and image collection; (b). VQA generation module; (c). Data filtering and search partitioning; (d). Visualization of problem type classification; (e). Visualization of search types.}
    \label{fig:data}
\end{figure}

Several search-based VQA datasets like LIVEVQA\cite{fu2025livevqa} and Infoseek\cite{chen2023can} have emerged as evaluation benchmarks. However, training datasets remain relatively small in scale and limited in source diversity, making it challenging to perform effective reasoning across broad aspects. For instance, MMSearch-R1\cite{wu2025mmsearch} constructed the multimodal search VQA dataset FactualVQA, but suffers from a single data source and lacks timely updates.
Therefore, we build a new search-based behavioral data Engine as shown at the top of Fig.~\ref{fig:data}, consisting of three main components: (1) Multimodal Information Acquisition; 
(2) VQA Question-Answer Pair Generation;
(3) Data Filtering and Classification and (4) Behavioral Metadata Generation. 

\textbf{Multimodal Data Acquisition.} To acquire comprehensive coverage and timeliness data, this study selected five global news platforms—including CNN and BBC—referencing LIVEVQA \cite{fu2025livevqa}. Unlike previous approaches, we employed a broader range of news channels to cover diverse domains. Specifically, we extracted topic information and corresponding images from each channel. During data collection, AD filtering were applied to preliminarily ensure the validity of multimodal information.

\textbf{VQA Pair Generation.} To ensure diversity and cost-effectiveness in QA generation, we deployed Qwen2.5-72B-Instruct\cite{bai2025qwen2}. Following LIVEVQA, we used events to determine Level 1 question types. Subsequently, Level 2 questions were generated from Level 1 questions, involving more complex reasoning processes with multi-level jump searches. Through this workflow, we preliminarily generated 8,000 image-based question-answer pairs.

\textbf{Data Filtering and Classification.} To ensure dataset quality and suitability for reinforcement learning, we performed data filtering and search type categorization. This process combined model-based and manual methods. Human reviewers examined data information, while models filtered out cases with unclear temporal context or mismatched images. Ultimately, as shown at the bottom of Fig.~\ref{fig:data}, we obtained 6,549 data entries, each containing elements such as question, image, and answer.

\textbf{Behavioral Metadata Generation}
We use the pre-trained Qwen2.5-VL-7B-Instruct\cite{bai2025qwen2} model to perform direct inference on our constructed training set. For each sample, we generate eight rollouts. If any of these eight rollouts produce a correct answer, we considered the model to have the potential to solve the problem with its internal knowledge and labeled the sample as \texttt{tool-free}. Conversely, if all eight rollouts failed to answer correctly, we determined that the model required external knowledge and labeled the sample as \texttt{tool-required}. While prior work \cite{wu2025mmsearch} employed a similar labeling procedure, their objective was limited to create a balanced dataset. Our goal, in contrast, is to generate an explicit signal that will be directly consumed by our process-oriented reward during training. Ultimately, this process resulted in 34.49\% of the samples in our constructed dataset being labeled as \texttt{tool-free}.

\begin{figure*}[t]
    \centering
    \includegraphics[width=0.8\textwidth]{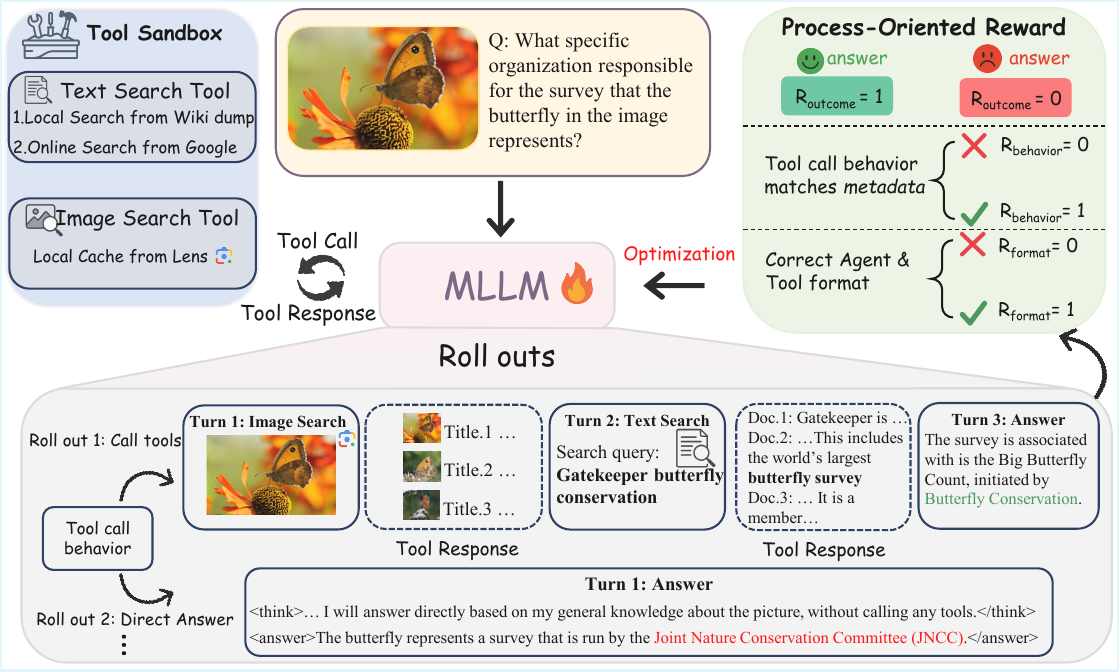} 
    \caption{The ProMMSearchAgent framework. The agent generates rollouts by deciding between direct answers or multi-step tool calls. The MLLM is optimized using GRPO. The reward signal is a composite of three components: outcome ($R_{\text{outcome}}$), behavior ($R_{\text{behavior}}$), and format ($R_{\text{format}}$) rewards.}
    \label{fig:overview}
\end{figure*}
\subsection{Decoupled Process-Oriented Reward}

As shown in the top-right corner of Fig.~\ref{fig:overview}, we propose a Decoupled Process-Oriented Reward for dealing with the logically contradictory between training signals and inhibits policy learning. We decompose the composite reward that isolates the final outcome from the process, allowing for clean, consistent credit assignment.

In detail, reward–action mismatch is exemplified by the reward structures used in prior work denoted as $R_{\text{conf}}$ . A specific instance is formulated as a weighted sum of a format score and an outcome score that is multiplicatively penalized by a tool-use action:
\begin{equation}
R_{\text{conf}} = (1-\alpha) \cdot R_{outcome} \cdot (1 - \mathbb{I}(P) \cdot p_t) + \alpha \cdot R_{format},
\end{equation}
where $R_{outcome}$ is the binary indicator for the final answer , $\mathbb{I}(P)$ is an indicator function that is $1$ if any tool was called, and $p_t$ is the "tool call penalty" factor .

This design creates a fundamental logical contradiction. For an instance where tool call is necessary ($M=\text{'tool-required'}$), the agent must call a tool ($\mathbb{I}(P)=1$) to achieve $R_{outcome}=1$. However, its potential reward is explicitly penalized by the $p_t$ term. This confounded signal inhibits the policy from learning a coherent strategy for when to use tools.

We solve this contradiction by first decomposing our reward, $R_{\text{decomp}}$, into its two primary components: 
\begin{equation}
R_{\text{decomp}} = \lambda_{\text{outcome}} R_{\text{outcome}} + \lambda_{\text{process}} R_{\text{process}}, 
\end{equation}
where $R_{\text{outcome}}$ is a sparse, standard reward for task completion. It is $1$ if the agent's final answer is judged correct against the ground truth, and $0$ otherwise.  $R_{\text{process}}$ is a dense reward for how the agent acts. It is itself a composite, further decomposed into:
\begin{equation}
R_{\text{process}} = w_{\text{behavior}} R_{\text{behavior}} + w_{\text{format}} R_{\text{format}}.
\end{equation}
$R_{\text{format}}$ ensures syntactic validity. It provides $+$ if the agent's trajectory adheres to the required ReAct-style syntax and issues a structurally valid tool call. This is similar in function to format scoring in prior works. $R_{\text{behavior}}$ provides a dense reward based only on the correctness of the agent's decision to act. It is an indicator function $R_{\text{behavior}} = \mathbb{I}(C_{\text{align}})$ that provides $+1$ if the agent's action $A$ aligns with the sample's metadata $M$. This alignment condition $C_{\text{align}}$ is met if:
\begin{equation}
(P = \text{'ToolCall'} \land M = \text{'req'}) \lor (P = \text{'Answer'} \land M = \text{'free'}),
\end{equation}
where $P$ is the agent's chosen policy (any tool call or a direct answer) and $M$ is the sample's metadata tag.

This hierarchical reward structure fundamentally resolves the baseline's logical flaw. Instead of punishing a necessary tool call , our $R_{\text{behavior}}$ explicitly and positively rewards the correct decision. This provides a clean, consistent, and dense signal for the RL optimizer to learn the primary agentic skill: correctly identifying its own knowledge boundary and acting accordingly.

\subsection{Sim-to-Real Decoupled Training Environment}
\textbf{Asynchronous RL Training.} We utilize the AgentLoop framework within veRL\cite{sheng2025hybridflow} for our reinforcement learning training, as it is an asynchronous agentic framework. This approach significantly improves data throughput in online reinforcement learning training. 

\textbf{Rollouts.}
As shown at the bottom of Fig.~\ref{fig:overview}, our agent's rollout trajectory follows the \texttt{ReAct}\cite{yao2023reactsynergizingreasoningacting} paradigm. The agent iteratively generates reasoning steps and interacts with the live environment using the provided tools until the model decides to produce a final answer or reaches our predefined maximum turn limit.

To enforce a structured reasoning process, the model's output is required to strictly adhere to specific XML tags:
\begin{itemize}
    \item All internal reasoning must be wrapped in \texttt{\textless{}think\textgreater{}...\textless{}/think\textgreater{}} tags.
    \item Tool calls must be wrapped in \texttt{\textless{}tool\_call\textgreater{}...\textless{}/tool\_call\textgreater{}}.
    \item The final answer must be wrapped within \texttt{\textless{}answer\textgreater{}...\textless{}/answer\textgreater{}} tags.
\end{itemize}

Following prior work, all responses returned by the tools are masked out to prevent information leakage and is thereby excluded from the loss computation.

\textbf{Tool Sandbox.} To ensure a stable interaction environment, expedite training, reduce operational costs, and enhance reproducibility, we implemented a tool sandbox, which is decoupled from the main training framework. This tool sandbox provides services for both text search and image search.

The text search tool is implemented with two distinct modes: a local search service and a web search service. For local search, we use the 2025 Wikipedia dump\cite{karpukhin-etal-2020-dense,10.5555/3524938.3525306}, which serves as the knowledge source and employ e5\cite{wang2024textembeddingsweaklysupervisedcontrastive} as the retriever. For web search, we build a live web search tool powered by the Serper API.

The image search tool is to perform image-to-image retrieval, using the original image from the VQA data sample as a query. It returns relevant images along with their associated textual information. Owing to the prohibitive cost and significant latency of live image-to-image retrieval, we opted to construct a pre-computed local cache before training and testing. We pre-processed all images in both the training and testing datasets using the Serper API's Google image search (Lens). The search API queries the web for content most relevant to the images in the dataset. We locally store the images and titles of the web url from the top-10 most relevant results. 
\section{Experiments}
\label{sec:exp}

\definecolor{tablegray}{gray}{0.85}

\begin{table*}[htbp]
\centering
\small 
\caption{Main results across five benchmarks. \textsuperscript{\dag} \;MMSearch-R1 results are obtained using its released model within our inference framework. To ensure fairness, the judge prompt from its original paper is adopted.
\textsuperscript{\ddag} \;DeepMMSearch-R1 and WebWatcher results are directly taken from their original papers.
\textsuperscript{\S} \;The LiveVQA result of WebWatcher corresponds to an earlier version of the LiveVQA benchmark and is therefore not strictly comparable to results obtained on the latest version used in this work.}
\begin{tabular*}{\textwidth}{@{\extracolsep{\fill}}lccccc}
\toprule
\textbf{Model} & \textbf{FVQA-test} & \textbf{InfoSeek} & \textbf{SimpleVQA} & \textbf{LiveVQA} & \textbf{MMSearch} \\
\midrule

\rowcolor{tablegray}
\multicolumn{6}{c}{\textit{Direct}} \\
GPT-4o           & 41.7  & 42.7  & 46.6  & 26.9  & 22.2  \\
Gemini 2.5 Pro   & 37.2  & 37.0  & 53.4  & 27.7  & 26.9  \\
Qwen2.5-VL-7B    & 20.9 & 23.9 & 30.4 & 8.3 & 7.2  \\
Qwen2.5-VL-32B   & 24.7 & 25.8 & 40.1 & 18.7 & 15.7  \\
Qwen2.5-VL-72B   & 30.3 & 37.2  & 50.5 & 8.2  & 13.5 \\
\midrule

\rowcolor{tablegray}
\multicolumn{6}{c}{\textit{ReAct Agent}} \\
Qwen2.5-VL-7B    & 34.2  & 28.3  & 35.8 & 10.7 & 21.1   \\
Qwen2.5-VL-32B   & 51.3  & 38.0  & 48.5 & 24.8 & 27.3   \\
MMSearch-R1\textsuperscript{\dag}    & 58.0    & 49.0    & 55.3 & 28.3 & 43.9 \\
DeepMMSearch-R1\textsuperscript{\ddag} & ---    & 47.5    & 55.9 & --- & --- \\
WebWatcher\textsuperscript{\ddag,\S} & ---    & ---    & 54.3 & 51.2 & 55.3 \\
ProMMSearchAgent  & 63.1    & 55.3   & 58.6 & 32.5 & 55.2 \\
\bottomrule
\end{tabular*}

\label{tab:combined-results}
\end{table*}
\subsection{Setup}
\textbf{Implementation Details.} We trained our model using the veRL\cite{sheng2025hybridflow}, adopting Qwen2.5-VL-7B model as the base model. For the reinforcement learning (RL) , we employ Group Relative Policy Optimization (GRPO)\cite{shao2024deepseekmath} as our optimization algorithm. We train the model for 2 epochs using a global batch size of 128, 8 rollouts, and a learning rate of $1 \times 10^{-6}$. The maximum number of rounds within each rollout is set to 10, which theoretically permits the model to make up to nine tool calls before generating the final response in the last round. For our Decomposed Composite Reward, the weighting coefficients $\lambda_{\text{outcome}}$ and $\lambda_{\text{process}}$ are set to 0.7 and 0.3, respectively. Within the Process Reward component, the weights for behavior and format, $w_{\text{behavior}}$ and $w_{\text{format}}$, are set to $2/3$ and $1/3$, respectively. During the training process, we employ Qwen2.5-72B-Instruct as the judge model to evaluate the outcome reward, $R_{\text{outcome}}$. Throughout both training and evaluation, all tool schemas strictly adhere to the JSON structure specified by the OpenAI tool calling standard. The model is likewise constrained to generate its tool-use requests in this JSON format.

\textbf{Tool Details.} 
During both training and evaluation, we equip the model with two distinct tools from our tool sandbox: a text search tool and an image search tool. All  tool interactions strictly adhere to the JSON schema specified by the OpenAI function-calling standard.

(1) Text search tool details. To ensure stable, reproducible training, prevent training experiment failures due to external tool unreliability, and mitigate the high costs, we employ the local search tool within our tool sandbox during the RL training phase. For the evaluation phase, we transition to the web search service within our tool sandbox. 

(2) Image search tool details.
For image search, we utilize the pre-computed image cache within the tool sandbox. The tool ultimately returns the top-k images most similar to the image of the VQA sample, along with their web titles.

\textbf{Benchmark.}
We evaluate our model's performance on five benchmarks spanning knowledge-intensive visual question answering (VQA) and multimodal retrieval-based reasoning: FVQA-test\cite{wu2025mmsearch}, Infoseek\cite{chen2023can}, SimpleVQA\cite{cheng2025simplevqa}, MMsearch\cite{jiang2024mmsearch}, LiveVQA\cite{fu2025livevqa}. Following MMSearch-R1, we utilize the full FVQA-test set. Specifically, for Infoseek, we selected a subset of 2,000 samples. For SimpleVQA, we selected 1,013 QA pairs containing only English. For MMSearch, we selected 171 samples that include images. For the LiveVQA benchmark, we note a significant discrepancy in the data's version history. While the MMSearch-R1 paper reports results on a 3,602-sample version, this version appears to be deprecated or no longer publicly accessible. We use the currently available 2,384-sample version.

\textbf{Baselines.}
We employ two distinct paradigms for inference. (1)The first is Direct Inference, where the model is not provided with any external callable tools and must rely solely on its internal parametric knowledge to generate answers. Under this paradigm, we benchmarked the performance of GPT-4o, Gemini-2.5 Pro, and the open-source Qwen2.5-VL series. (2)The second is the ReAct\cite{yao2023reactsynergizingreasoningacting} Agent paradigm, an agentic framework where the model dynamically interleaves reasoning and action to solve tasks. Instead of answering directly, the model iteratively generates a reasoning trace ('Thought') to plan its next step, executes an 'Action' (e.g., invoking a search tool), and then 'Observes' the tool's response. This "Thought-Action-Observation" loop is repeated until the model gathers sufficient information to formulate the final answer. Under the ReAct agent paradigm, we compare our model against several baselines, including the Qwen2.5-VL-7B, MMSearch-R1, DeepMMSearch-R1, and WebWatcher. For DeepMMSearch-R1, we report the results directly from their paper, as its model weights have not been publicly released. Similarly, for WebWatcher, we also cite its paper-reported results. Its methodology relies on several tools (e.g., a Python interpreter, web-visit) that are not implemented in our tool sandbox, precluding a direct and fair comparison. 

\textbf{Datasets.}
Our training corpus consists exclusively of the dataset described in Sec. \ref{sec:data}, comprising a total of 6,549 samples. Within this training set, the proportion of tool-free instances is 34.49\%.

\textbf{Metrics.} We evaluate our models using two primary metrics: Accuracy (Acc.) and Tool Call Ratio (TR). For Accuracy, we adopt an LLM-as-judge framework to evaluate the correctness of the model's predictions. Specifically, we employ Qwen2.5-72B-Instruct as our judge model. The judge is provided with the input question, the ground truth, and the model's prediction to determine the correctness of the output. This is our primary metric for performance in the main results table. For Tool Call Ratio, we measure the percentage of test samples for which the model used at least one tool. We note that TR is highly sensitive to the agent's framework, tool format, and prompt style. As such, comparing TR values across different baselines is 
unfair, as their tool formats and reported TRs are not comparable. Therefore, we use TR as a key diagnostic metric, primarily in Sec.~\ref{sec:discuss}. In this section, all models share our identical framework, making the TR comparison a fair and powerful tool for analyzing the agent's decision-making policy.

\subsection{Main Results}

Table~\ref{tab:combined-results} shows the performance under the Direct and ReAct Agent paradigms.

Under the \emph{Direct} setting, all models, including strong closed-source systems (GPT-4o, Gemini 2.5 Pro) and large open-source models (Qwen2.5-VL-72B), perform poorly on these knowledge-intensive benchmarks. This confirms that the tasks are designed to require information beyond parametric knowledge and motivates tool-augmented agents.

Under the \emph{ReAct Agent} setting, enabling Qwen2.5-VL models to use tools leads to clear gains over direct inference, but ReAct prompting still leaves a large performance gap. Our ProMMSearchAgent achieves the best results on all five benchmarks, surpassing MMSearch-R1 by +5.1, +6.3, +3.3, +4.2, and +11.3 Acc. on FVQA-test, InfoSeek, SimpleVQA, LiveVQA, and MMSearch, respectively. Compared with a ReAct-style Qwen2.5-VL-32B agent, our 7B-based agent attains even larger improvements , indicating that our training recipe, rather than model size, is the key factor.

\begin{figure*}[t]
    \centering
\includegraphics[width=0.9\textwidth]{case1_cropped.pdf} 
    \caption{Case Comparison of Complex Multi-Step Reasoning.
The baseline model (left) employs a deviant search-reasoning strategy, leading to an incorrect answer. Our model (right) first locates the key entity via image search, then uses this cue to execute a precise text search, reasoning successfully to the correct answer.}
    \label{fig:case1}
\end{figure*}

Despite relying on a more restricted toolset, ProMMSearchAgent is competitive with WebWatcher: it slightly improves upon WebWatcher on SimpleVQA and essentially matches performance on MMSearch. WebWatcher reports higher numbers on LiveVQA, but those are based on an earlier version of the benchmark and are not directly comparable. Overall, the consistent gains across all five datasets---covering factual VQA, web-grounded QA, and multimodal retrieval---suggest that our agent has learned a generalizable and effective tool-use policy rather than overfitting to a specific benchmark.

\subsection{Discussions and Analysis}
\label{sec:discuss}
In this section, we delve into the sources of our model's performance advantage. We designate our full model, ProMMSearchAgent, as the primary baseline. All ablated model variants were trained using the identical training data and framework to ensure a fair comparison.

To rigorously validate the design of our decomposed reward function ($R_{\text{decomp}}$), we compare our proposed \texttt{ProMMSearchAgent} against four distinct reward configurations. Each configuration modifies specific components or weights of the reward function while keeping the training data and environment constant.

\textbf{ConfReward:} This setting replicates the "tool call penalty" logic used in MMSearch-R1. The reward is composed of 90\% Outcome and 10\% Format. However, if the agent performs \textit{any} tool call during the trajectory, a multiplicative penalty of 0.9 is applied to the outcome reward. 

    \textbf{NoBehaviorReward:} This setting removes the behavior reward. The reward consists solely of 90\% outcome reward and 10\% format reward.

    \textbf{NaiveBehaviorReward:} This setting tests the importance of our metadata-driven alignment. It maintains the 7:2:1 ratio, but replaces the behavior reward function with a naive signal that simply rewards $+1$ for \textit{any} tool call, regardless of necessity. 

    \textbf{HigherBehaviorReward:} To evaluate hyperparameter robustness, we alter the component weights of our proposed $R_{\text{decomp}}$ to a 6:3:1 ratio.

\begin{table*}[htbp] 
\centering
\caption{Ablation study of reward function components. Acc. denotes Accuracy and TR denotes Tool Call Ratio (\%).}
\label{tab:ablation_reward}
\small 
\setlength{\tabcolsep}{0pt} 
\begin{tabular*}{\textwidth}{@{\extracolsep{\fill}} cl cc cc cc cc cc}
\toprule
& & \multicolumn{2}{c}{FVQA-test} & \multicolumn{2}{c}{InfoSeek} & \multicolumn{2}{c}{SimpleVQA} & \multicolumn{2}{c}{LiveVQA} & \multicolumn{2}{c}{MMSearch} \\
\cmidrule(lr){3-4} \cmidrule(lr){5-6} \cmidrule(lr){7-8} \cmidrule(lr){9-10} \cmidrule(lr){11-12}
\# & \textbf{Model Configuration} & Acc. & TR & Acc. & TR & Acc. & TR & Acc. & TR & Acc. & TR \\
\midrule
0 & \texttt{Qwen2.5-VL-7B} & 34.2 & 84.4 & 28.3 & 86.1 & 35.8 & 80.5 & 10.7 & 71.4 & 21.1 & 86.2 \\
\midrule
\rowcolor{gray!10}
1 & \textbf{ProMMSearchAgent} & \textbf{62.4} & 88.1 & \textbf{55.3} & 90.3 & \textbf{58.6} & 74.7 & \textbf{32.5} & 94.0 & \textbf{55.2} & 96.7 \\
2 & \texttt{ConfReward}  & 56.9 & 83.4 & 49.7 & 83.2 & 54.7 & 63.4 & 29.1 & 88.4 & 48.2 & 91.2 \\
3 & \texttt{NoBehaviorReward}  & 61.6 & 99.9 & 52.1 & 100 & 58.5 & 99.0 & 32.3 & 100 & 58.5 & 100 \\
4 & \texttt{NaiveBehaviorReward} & 58.8 & 99.8 & 50.3 & 100 & 56.7 & 98.0 & 31.2 & 100 & 46.8 & 100 \\
5 & \texttt{HigherBehaviorReward} & 57.6 & 88.2 & 51.3 & 94.8 & 55.1 & 71.4 & 30.3 & 95.0 & 53.0 & 96.3 \\
\bottomrule
\end{tabular*}
\end{table*}

\textbf{Necessity of RL Training.}
The results in Row~0, which represent the base Qwen2.5-VL-7B model operating via prompting alone, are highly revealing. The base model already achieves a high tool call ratio (TR), yet its accuracy is low (28.3\%). The main issues lie in how tools are used. The agent frequently mis-selects between image and text search, formulates suboptimal queries, and fails to reliably incorporate retrieved evidence into its reasoning. This highlights a key role of our reinforcement learning stage: rather than simply encouraging more tool calls, it teaches the agent to reason about tools and operate them effectively. 

\textbf{Limitations of Penalty-Based Rewards.}
Row~2 (\texttt{ConfReward}) applies the tool call penalty from \texttt{MMSearch-R1}. This term successfully suppresses TR on simpler datasets (e.g., 63.4\% vs.\ 74.7\% on SimpleVQA) but also unduly discourages search on difficult benchmarks. On LiveVQA and MMSearch, its TR is consistently lower than our full model (88.4\% vs.\ 94.0\%), resulting in lower accuracy. This suggests that penalty-only signals can lead to a ``tool-shy'' policy that under-utilizes search for genuinely knowledge-intensive queries.

\textbf{Effect of Meta data Driven Behavior Reward.} Row~3 removes the process reward, retaining only the outcome term. In this setting, the agent's policy collapses to near-constant tool invocation, with TR approaching 100\% on all datasets. Lacking a guiding process signal, the model resorts to call tools as much as possible to maximize the outcome reward. While the accuracy does not degrade catastrophically, the policy is highly inefficient and fails to identify tool-free instances, rendering it clearly inferior to our metadata-guided design.

The results for Row 4 (\texttt{NaiveBehaviorReward}) reinforces that naively rewarding all tool usage is detrimental. Similar to \texttt{NoBehaviorReward}, this strategy inflates the TR to nearly 100\%. We attribute this to a classic case of reward hacking. It defaults to a degenerate policy of $\sim$100\% Search Ratio to trivially maximize the dense process reward, disregarding the actual utility of the search. This behavior, however, ignores the actual utility of the search action, leading to the observed collapse in task performance.

In stark contrast, ours (Row~1) does not merely encourage exploration; it selectively aligns tool invocation with the model's own internal knowledge boundary. This demonstrates that our design effectively enhances both the effectiveness and the selectivity of tool utilization.

\textbf{Adaptive behavior and robustness.}
Our full model (Row~1) exhibits adaptive behavior. Compared to the base agent (Row~0) and the penalty-based Reward (Row~2), it strategically reduces TR on simpler benchmarks (e.g., SimpleVQA: 80.5\% $\rightarrow$ 74.7\%) while increasing TR on more difficult ones (e.g., LiveVQA: 71.4\% $\rightarrow$ 94.0\%). Row~5 further demonstrates that this adaptive behavior and the resulting accuracy are stable under moderate hyperparameter changes, indicating a robust learned policy rather than one hyperspecific to a single configuration.

\textbf{Metadata Analysis}
To ensure these self-generated labels reflect genuine knowledge gaps rather than mere reasoning failures, we conducted a manual audit alongside a GPT-4o verification on 200 randomly sampled "tool-required" instances. The audit confirmed that 78.5\% of these cases genuinely required external factual knowledge to resolve visual ambiguities. Furthermore, for the minority of cases where the model possessed parametric knowledge but failed to reason correctly, the search tool serves as a "Corrective Verifier." By retrieving explicit context, the agent grounds its visual reasoning, transforming potential hallucinations into verified multimodal alignments.


\textbf{Analysis of Query Quality and Transferability.}
To verify the intrinsic quality of generated queries and the efficacy of our Sim-to-Real paradigm, we conducted a controlled "search-backbone-swap" analysis. Specifically, for 246 instances from the aforementioned 371-sample subset where the agent triggered a text search, we kept the model-generated query strings identical but replaced the live Google Search API's retrieval results with responses from the local sandbox. The model then performed its final reasoning based on this external, unseen evidence. Using the exact same queries, the live Google Search API achieved an accuracy of 64.1\%, significantly surpassing the local sandbox (56.1\%). This performance leap provides empirical proof that our agent masters an engine-agnostic search logic rather than overfitting to the local retriever. Qualitatively, we observed a distinct query evolution: early checkpoints often generated generic queries (e.g., "Who is the person in the image?"), but these were rapidly pruned during training in favor of specific, entity-centric formulations. This evolution demonstrates that our process-oriented reward effectively teaches the agent to distill pivotal visual entities into precise search terms, facilitating a seamless and robust zero-shot transfer from the local training environment to the live web.

\begin{table}[tb]
\caption{Performance and cost-efficiency trade-off, measured after one epoch of training. Accuracy is averaged across all 5 test benchmarks.}
\centering
\begin{tabular}{l|c|c|c}
\toprule
\textbf{Training Tools} & \textbf{Avg. Acc.} & \textbf{API Cost} & \textbf{Tool Call Time (s)} \\
\midrule
\rowcolor{gray!10}
\texttt{Online} & \textbf{50.71\%} & \$0.50 / 1k calls & 6017.81 s \\
\texttt{Local} & 48.83\% & \textbf{Virtually \$0.00} & \textbf{457.30 s} \\

\end{tabular}%

\label{tab:local&online}
\end{table}
\subsection{Local/Online Training Environment}

A core thesis of our work is that a robust process-oriented reward ($R_{\text{decomp}}$) facilitates the acquisition of an invariant search policy. This enables agents to be trained effectively in a deterministic, local-first environment, effectively bridging the gap between static simulation and the volatile live web.

To validate this claim, we performed a direct comparison. We trained two models for a single epoch with identical settings, differing only in their tool backend: 1) \texttt{LocalSearch} (1-epoch): Trained on our local Wiki corpus; 2) \texttt{OnlineSearch} (1-epoch): Trained using the live Serper Google Search API. Table \ref{tab:local&online} presents the trade-off between accuracy, cost, and speed. We report the average accuracy across all five test benchmarks.

As shown in Table \ref{tab:local&online}, while \texttt{OnlineSearch} maintains a marginal 1.88\% accuracy advantage,  \texttt{LocalSearch} retains over 96\% of the online-trained performance despite never accessing the live web. This confirms the agent masters an invariant, generalizable search logic rather than memorizing engine-specific results.

This minor accuracy trade-off yields massive gains: local training is 13.1$\times$ faster (457.3s vs. 6017.8s), eliminates prohibitive API costs, and provides a deterministic optimization landscape. Thus, our Sim-to-Real paradigm is not merely a cost-saving measure, but a scientifically superior and reproducible pipeline.

\section{Conclusion}
We presented ProMMSearchAgent, a novel framework that resolves the critical reward-action mismatch and environmental bottlenecks in training multimodal search agents. By proposing an introspective process-oriented reward, we explicitly teach the agent to recognize its visual and factual knowledge boundaries. Coupling this dense supervision with a deterministic Sim-to-Real sandbox enables a fully reproducible, low-cost training paradigm. Ultimately, our agent learns an invariant search logic that generalizes zero-shot to live web APIs, achieving state-of-the-art performance on knowledge-intensive VQA benchmarks while demonstrating genuine meta-cognitive calibration in its adaptive tool use.

%
%
\bibliographystyle{splncs04}
\bibliography{main}
\appendix
\section{System Prompt}
\label{Sysprompt}
Here is the system prompt utilized during both training and inference. It incorporates detailed tool specifications and outlines the ReAct agent's operational workflow.
\begin{tcolorbox}[
    enhanced,
    colback=promptbody,
    colframe=promptheader,
    coltitle=white,
    title=\textbf{System Prompt for ReAct Agent},
    fonttitle=\bfseries\large,
    arc=3mm,
    boxrule=0.5mm,
    left=2mm, right=2mm, top=2mm, bottom=2mm,
    standard jigsaw,
    opacityback=1,
    breakable
]
    You are a helpful assistant. You can call functions to assist with the user query.
    
    \textbf{Important:} You must call only one function at a time. After each function call, wait for the execution result before making the next function call if needed.
    
    \vspace{0.5em}
    \textbf{\# Tools}
    
    You may call one or more functions to assist with the user query.
    
    You are provided with function signatures within \texttt{<tools></tools>} XML tags:
    
    \vspace{0.5em}
    \begin{small}
    \texttt{<tools>}
    
    \texttt{\{"type": "function", "function": \{"name": "\textbf{search}", "description": "Searches the web for relevant information based on the given query.", "parameters": \{"type": "object", "properties": \{"query\_list": \{"type": "array", "description": "A list of complete textual queries. Each query must clearly state the topic without referring to images."\}\}, "required": ["query\_list"]\}\}\}}
    
    \vspace{0.3em}
    \texttt{\{"type": "function", "function": \{"name": "\textbf{web\_image\_to\_image\_search}", "description": "**IMPORTANT: This tool can only be called once.** Searches for relevant images based on the original image using web search."\}\}}
    
    \texttt{</tools>}
    \end{small}
    \vspace{0.5em}
    
    For each function call, return a json object with function name and arguments within \texttt{<tool\_call></tool\_call>} XML tags:
    
    \vspace{0.5em}
    \texttt{<tool\_call>}
    
    \texttt{\{"name": <function-name>, "arguments": <args-json-object>\}}
    
    \texttt{</tool\_call>}

    \vspace{1em}
    \hrule
    \vspace{1em}

    \textbf{Workflow and Output Format}
    
    You must follow these steps in order. In every conversation turn, you start from Step 1.
    
    \vspace{0.5em}
    \textbf{Step 1: Think (Plan)}
    \begin{itemize}[leftmargin=*, nosep]
        \item \textbf{This is the starting point for every turn.}
        \item Analyze the user's query and all available information (including previous observations) carefully.
        \item Formulate a plan. Your plan must decide on \textbf{one} of two courses of action:
        \begin{enumerate}[nosep]
            \item \textbf{Call a tool:} If you need more information.
            \item \textbf{Provide a final answer:} If you have sufficient information.
        \end{enumerate}
        \item Your entire reasoning process must be enclosed in \texttt{<think>...</think>} tags.
    \end{itemize}
    
    \vspace{0.5em}
    \textbf{Step 2: Act (Tool Call)}
    \begin{itemize}[leftmargin=*, nosep]
        \item \textbf{Execute this step ONLY if your Step 1 plan was to call a tool.}
        \item Call the \textbf{one single tool} decided upon in your plan.
        \item The tool call must be enclosed in \texttt{<tool\_call>...</tool\_call>} tags.
        \item \textbf{Important: If you call a tool, you must STOP and wait for the observation. Do NOT proceed to Step 4.}
    \end{itemize}
    
    \vspace{0.5em}
    \textbf{Step 3: Observe (Tool Output)}
    \begin{itemize}[leftmargin=*, nosep]
        \item \textbf{You will only enter this step after a tool call.}
        \item You will receive the tool's output (observation).
        \item After receiving the output, you \textbf{MUST} go back to \textbf{Step 1 (Think)} to analyze the new information and decide the next step (e.g., call another tool or provide the final answer).
    \end{itemize}
    
    \vspace{0.5em}
    \textbf{Step 4: Answer (Final Response)}
    \begin{itemize}[leftmargin=*, nosep]
        \item \textbf{Execute this step ONLY if your Step 1 plan was to provide a final answer.}
        \item In your \textbf{Step 1 Think block}, you must have already synthesized all information and planned the content of your response.
        \item Formulate the \textbf{comprehensive, detailed, and user-facing final answer} based on that plan.
        \item Your final answer must be enclosed in \texttt{<answer>...</answer>} tags.
    \end{itemize}
    
    \vspace{0.5em}
    Here is the question and image: \\
    \texttt{<image>}
\end{tcolorbox}

\begin{tcolorbox}[
    enhanced,
    colback=promptbody,
    colframe=promptheader,
    coltitle=white,
    title=\textbf{Prompt for Direct Answer},
    fonttitle=\bfseries\large,
    arc=3mm,
    boxrule=0.5mm,
    left=2mm, right=2mm, top=2mm, bottom=2mm,
    standard jigsaw,
    opacityback=1,
    breakable
]
Based on the image, answer the question. Here is the question and image:
\\
    \texttt{<image>}
\end{tcolorbox}

\section{Judge Prompt}
\label{Judge Prompt}
We employ both Qwen2.5-72B and Qwen3-14B as our judge models. The specific prompt utilized to evaluate prediction accuracy is presented below:
\begin{tcolorbox}[
    enhanced,
    colback=promptbody,
    colframe=promptheader,
    coltitle=white,
    title=\textbf{Judge Prompt},
    fonttitle=\bfseries\large,
    arc=3mm,
    boxrule=0.5mm,
    left=2mm, right=2mm, top=2mm, bottom=2mm,
    standard jigsaw,
    opacityback=1,
    breakable
]
    Judge whether the following [response] to [question] is correct or not based on the precise and unambiguous [correct\_answer] below.
    
    \vspace{0.5em}
    [question]: \textbf{\{question\}}
    
    \vspace{0.5em}
    [response]: \textbf{\{response\}}
    
    \vspace{0.5em}
    Your judgement must be in the format and criteria specified below:
    
    \vspace{0.5em}
    extracted\_final\_answer: The final exact answer extracted from the [response]. Put the extracted answer as 'None' if there is no exact, final answer to extract from the response.
    
    \vspace{0.5em}
    [correct\_answer]: \textbf{\{correct\_answer\}}
    
    \vspace{0.5em}
    reasoning: Explain why the extracted\_final\_answer is correct or incorrect based on [correct\_answer], focusing only on if there are meaningful differences between [correct\_answer] and the extracted\_final\_answer. Do not comment on any background to the problem, do not attempt to solve the problem, do not argue for any answer different than [correct\_answer], focus only on whether the answers match.
    
    \vspace{0.5em}
    correct: Answer 'yes' if extracted\_final\_answer matches the [correct\_answer] given above, or is within a small margin of error for numerical problems. Answer 'no' otherwise, i.e. if there if there is any inconsistency, ambiguity, non-equivalency, or if the extracted answer is incorrect.
    
    \vspace{0.5em}
    confidence: The extracted confidence score between 0|\%| and 100|\%| from [response]. Put 100 if there is no confidence score available.
\end{tcolorbox}

\section{Implementation Details}
\subsection{Local Knowledge 
Base Construction Details }
\label{subsection:kknowledge base}
To ensure the reproducibility of our local training environment and the quality of the retrieved knowledge, we constructed a static local knowledge base derived from the English Wikipedia. The construction process involves three stages: Data Filtering, Hierarchical Cleaning, and Context-Aware Chunking. 

\textbf{Data Source and Filtering.} We utilize the standard Wikipedia dump (enwiki-20251020-pages-articles-multistream.xml.bz2). To guarantee the retrieval of high-quality factual information, we apply a strict filtering pipeline. A page $P$ is retained only if it satisfies the following conditions:\begin{itemize}\item Namespace Check: The page must belong to the main namespace ($NS=0$). Meta-pages such as Category, Template, Talk, and File are discarded.\item Disambiguation Filter: Pages explicitly marked as disambiguation pages are excluded to prevent the retrieval of semantic ambiguity.\item Redirect Filter: We exclude redirect pages to ensure that the knowledge base consists solely of canonical content.\end{itemize}

\textbf{Hierarchical Context Injection.} A common issue in RAG systems is the loss of global context when a document is split into chunks. To address this, we implement a section-aware processing strategy. For a specific section $S$ within a page having title $T$, we explicitly inject the hierarchical context into the raw text content $C_{raw}$. The effective content $C_{eff}$ is formulated as:$$C_{\text{eff}} = f_{\text{clean}}\left( \text{``} T \text{ --- } S \text{''} \oplus \texttt{"\textbackslash n"} \oplus C_{\text{raw}} \right)$$where $\oplus$ denotes string concatenation, and $f_{clean}(\cdot)$ represents the cleaning function that removes Wiki-markup (e.g., [[links]], {{templates}}, \textless ref\textgreater tags\textless /ref\textgreater) using regular expressions. This ensures that every retrieved chunk carries its topic and section information.

\textbf{Sentence-Aware Chunking and Deduplication.} 
\label{subsection:index}
Instead of fixed-size token truncation which may break semantic coherence, we employ a sentence-aware chunking algorithm. The text is first segmented into sentences using punctuation delimiters (e.g., ., ?, !). We accumulate sentences into a chunk $K$ until its length approaches a maximum threshold $L_{max}$.The constraints for a valid chunk $K$ are defined as:$$L_{min} \le \text{len}(K) \le L_{max}$$In our implementation, we set $L_{max} = 700$ characters to reduce the total number of chunks while maintaining sufficient context, and $L_{min} = 220$ characters to filter out uninformative fragments. Finally, to remove redundant information, we apply MD5-based deduplication by computing the hash of the normalized text (lowercased and whitespace-collapsed).

\subsection{Dense Indexing Implementation}

Following the construction of the local text corpus, we build a dense vector index to enable efficient semantic retrieval. We employ the \textbf{E5} (EmbEddings from bidirEcTional Encoder rEpresentations) model family as our backbone encoder. The indexing pipeline consists of three strictly defined stages: asymmetric formatting, mean-pooled encoding, and normalized FAISS indexing.

\textbf{Asymmetric Input Formatting.}
E5 models are trained with an asymmetric structure that distinguishes between queries and passages. To align with the model's pre-training objective, we apply a specific prefix to the input text. For every document chunk $d_i$ in our processed corpus, we prepend the context identifier \texttt{"passage: "} prior to tokenization:
\begin{equation}
    x_i = \texttt{"passage: "} \oplus d_i
\end{equation}
where $\oplus$ denotes string concatenation. This prefix guides the attention mechanism to generate embeddings optimized for passive retrieval, distinguishing them from query representations (which are prefixed with \texttt{"query: "} during the inference phase).

\textbf{Encoding and Pooling Strategy.}
We utilize the Transformer-based encoder to process the formatted input $x_i$. The maximum sequence length is set to $L=180$ tokens to fully cover the chunk size defined in the previous section. Let $H_i \in \mathbb{R}^{L \times h}$ be the last hidden states output by the encoder, and $M_i \in \{0, 1\}^L$ be the attention mask where $1$ indicates a valid token. We apply \textbf{Mean Pooling} to derive the raw vector representation $v_{\text{raw}}$:
\begin{equation}
    v_{\text{raw}} = \frac{\sum_{j=1}^{L} (H_{i,j} \cdot M_{i,j})}{\sum_{j=1}^{L} M_{i,j}}
\end{equation}
This strategy averages the token embeddings to capture global semantic information, which differs from the \texttt{[CLS]} token pooling used in standard BERT models.

\textbf{L2-Normalization and Indexing.}
To ensure that the Inner Product metric used during retrieval is mathematically equivalent to Cosine Similarity, we apply $\ell_2$-normalization to the pooled vectors:
\begin{equation}
    v_i = \frac{v_{\text{raw}}}{\|v_{\text{raw}}\|_2}
\end{equation}
The normalized embeddings $\{v_i\}_{i=1}^N$ are stored using the \textbf{FAISS} library. We construct a \texttt{Flat} index utilizing the Inner Product metric (\texttt{METRIC\_INNER\_PRODUCT}) for exact nearest neighbor search. To handle the large-scale indexing efficiently, we implement parallel inference using \texttt{DataParallel} across multiple GPUs with half-precision (FP16) optimization.

\subsection{Retrieval Inference}
\label{subsec:retrieval}
\textbf{Asymmetric Query Processing.}
Consistent with the indexing phase, we strictly adhere to the asymmetric instruction format of the E5 model during inference. For an incoming user query $q$, we prepend the query-specific prefix before tokenization:
\begin{equation}
    q_{\text{input}} = \texttt{"query: "} \oplus q
\end{equation}
This input is processed by the same encoder $f_\theta$ used for indexing. We apply the identical Mean Pooling and $\ell_2$-normalization strategies to obtain the query embedding $v_q$:
\begin{equation}
    v_q = \frac{\text{MeanPool}(f_\theta(q_{\text{input}}))}{\|\text{MeanPool}(f_\theta(q_{\text{input}}))\|_2}
\end{equation}
This ensures that the vector space alignment between the query and the pre-indexed passages is preserved.

\textbf{Fast Retrieval via GPU-Accelerated Search.}
We utilize the \textbf{FAISS} library to perform similarity search. To minimize latency during the Reinforcement Learning (RL) training loop, the pre-built index is loaded into GPU memory with half-precision (FP16) optimizations enabled (\texttt{GpuMultipleClonerOptions}).
For a given query vector $v_q$, the retriever returns the top-$k$ documents $D_{topk}$ based on the inner product score (equivalent to cosine similarity due to normalization):
\begin{equation}
    D_{topk} = \operatorname*{arg\,topk}_{d_i \in \mathcal{C}} \langle v_q, v_{d_i} \rangle
\end{equation}
where $v_{d_i}$ represents the pre-computed embedding of document $d_i$ in the corpus $\mathcal{C}$.

\section{Training Hyperparameters}

We provide the detailed hyperparameter configuration used for training ProMMSearchAgent in Table ~\ref{tab:hyperparams}. The model was trained using the Group Relative Policy Optimization (GRPO) algorithm on the \texttt{verl} framework.

\begin{table}[!b]
    \centering
    \small
    \renewcommand{\arraystretch}{1.2}

    \caption{Detailed Hyperparameters for Training.}
    \begin{tabular}{l|c}
        \toprule
        \textbf{Hyperparameter} & \textbf{Value} \\
        \midrule
        \multicolumn{2}{c}{\textit{Optimization Configuration}} \\
        \midrule
        Base Model & Qwen2.5-VL-7B \\
        Optimization Algorithm & GRPO \\
        Learning Rate & $1 \times 10^{-6}$ \\
        Global Batch Size & 128 \\
        PPO Mini-Batch Size & 128 \\
        Total Epochs & 2 \\
        Optimizer Offload & True \\
        Param Offload & True \\
        \midrule
        \multicolumn{2}{c}{\textit{Rollout \& GRPO Settings}} \\
        \midrule
        Number of Rollouts per Prompt & 8 \\
        KL Coefficient ($\beta$) & 0.0 \\
        Advantage Estimator & GRPO \\
        Entropy Coefficient & 0.0 \\
        Rollout Engine & SGLang (Async) \\
        \midrule
        \multicolumn{2}{c}{\textit{Sequence \& Agent Settings}} \\
        \midrule
        Max Prompt Length & 8192 \\
        Max Response Length & 16384 \\
        Max Assistant Turns & 10 \\
        Max User Turns & 10 \\
        Max Tool Response Length & 4096 \\
        Hardware Resources & $8 \times$ GPUs (Single Node) \\
        \bottomrule
    \end{tabular}

    \label{tab:hyperparams}
\end{table}

\section{Multimodal Search Tools}
\subsection{Tool  Definition}
All tools are defined in strict accordance with the OpenAI function tool schema. The specific tool prompts, provided in Section ~\ref{Sysprompt}, stipulate both the invocation syntax and the semantic capabilities of each tool. The model is instructed to execute tool calls within the \texttt{<tool\_call>}...\texttt{</tool\_call>} block. Specifically, our tool definitions are structured as follows
\begin{tcolorbox}[
    enhanced,
    colback=promptbody,
    colframe=promptheader,
    coltitle=white,
    title=\textbf{Tool: Search},
    fonttitle=\bfseries\large,
    arc=3mm,
    boxrule=0.5mm,
    left=2mm, right=2mm, top=2mm, bottom=2mm,
    standard jigsaw,
    opacityback=1,
    breakable
]
 \textbf{Description}\par
Searches the web for relevant information based on the given query.

\medskip

\textbf{Arguments}
\begin{itemize}
    \item \texttt{query\_list} (array of strings):  
    A list of complete textual queries. Each query must clearly state the topic without referring to images.
\end{itemize}
\end{tcolorbox}
\begin{tcolorbox}[
    enhanced,
    colback=promptbody,
    colframe=promptheader,
    coltitle=white,
    title=\textbf{Tool: ImageSearch},
    fonttitle=\bfseries\large,
    arc=3mm,
    boxrule=0.5mm,
    left=2mm, right=2mm, top=2mm, bottom=2mm,
    standard jigsaw,
    opacityback=1,
    breakable
]
 \textbf{Description}\par
Searches for relevant images based on the original image using web search.

\medskip

\textbf{Arguments}
\begin{itemize}
    \item \textbf{None}
\end{itemize}
\end{tcolorbox}

\subsection{Tool Details}
\textbf{Local Search Tool.}
For the local retrieval setup, we use Wiki25 as the corpus and E5 as the retriever. We refer to Section ~\ref{subsection:kknowledge base} for the database construction, Section ~\ref{subsection:index} for the index construction, and Section ~\ref{subsec:retrieval} for the retrieval mechanism. In the training stage, the tool is configured to return the top-3 most relevant documents.

\textbf{Web Search Tool} For the web search setup, we utilize the Serper API—which provides an interface to Google Search—as our online search service. In the testing stage, this tool returns the top-5 most relevant results for the given query. We emphasize that no web scraper or summarization model is employed. The model is directly provided with the raw Serper API Search output, which consists of the 'title', 'link', and 'snippet' for each result. To minimize the overhead associated with the web search tool, we implemented a Redis-based caching layer. When the tool sandbox encounters a query identical to a previously processed one, it retrieves the result directly from the cache, effectively bypassing the execution of a live web search.

\textbf{Image Search Tool} For image search set up, we utilize the Google Lens service via the Serper API to obtain the top-$k$ visually similar images and their corresponding webpage titles. To address the high latency and instability of live API calls during training, we implemented a deterministic offline cache. Specifically, we performed a one-time pre-computation for all images in the training and test sets. The search results are serialized locally and indexed by the unique \texttt{Data ID} of each sample. During the agent's execution, the image search tool functions as an $O(1)$ file lookup that returns these pre-cached results instantaneously, thereby ensuring a zero-cost and fully reproducible experimental environment. At runtime, when the model calls the image search tool, the top-3 cached results will be returned to the model.

\section{Case Study}

\definecolor{promptheader}{RGB}{70, 130, 180} 
\definecolor{promptbody}{RGB}{245, 248, 250}   
\definecolor{thinkcolor}{RGB}{105, 105, 105}   
\definecolor{answercolor}{RGB}{0, 100, 0}     
\begin{tcolorbox}[
    enhanced,
    colback=promptbody,
    colframe=promptheader,
    coltitle=white,
    title=\textbf{\faRobot\ Case 1 from FVQA-test},
    fonttitle=\bfseries\large,
    arc=3mm,
    boxrule=0.5mm,
    left=3mm, right=3mm, top=3mm, bottom=3mm, 
    standard jigsaw,
    opacityback=1,
    breakable
]
    \textbf{\faQuestionCircle\ Question:} Where is this skyline located?
    
    \begin{center}
        \includegraphics[width=0.85\linewidth, height=4cm, keepaspectratio]{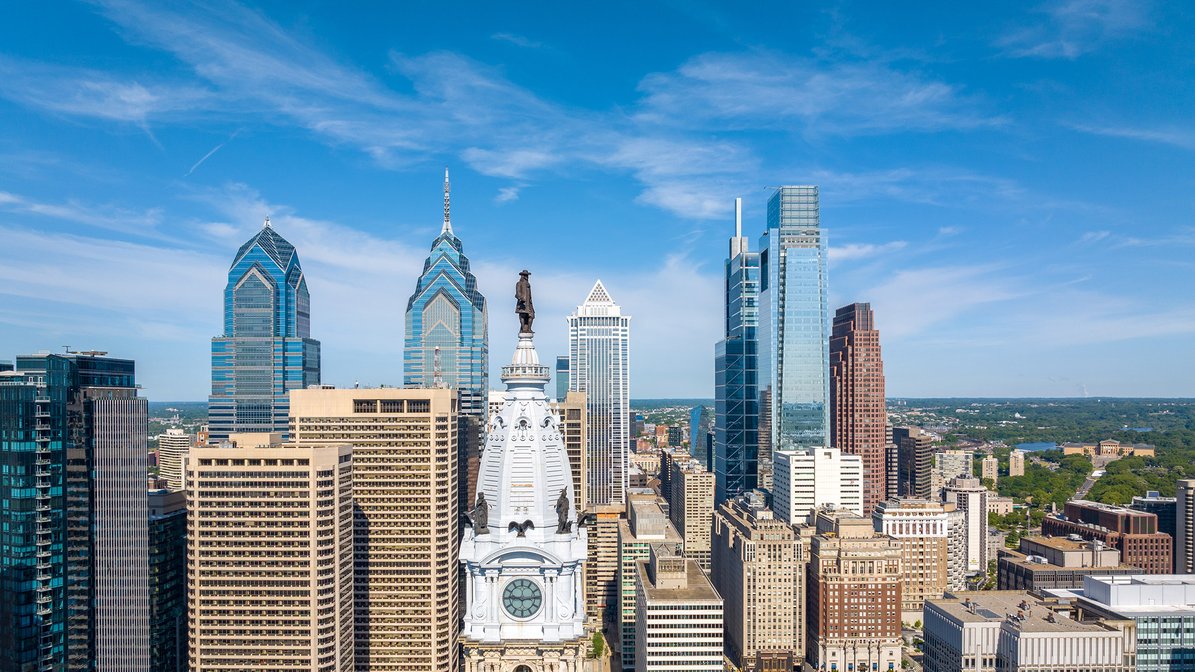}
    \end{center}
    
    \textbf{\faCommentDots\ Model Response:}
    
    \vspace{0.5em}
    {\color{thinkcolor}
    \texttt{<think>}
    The skyline in the image appears to be of Philadelphia, given the iconic City Hall clock tower. No additional information is necessary to identify this location directly from the image.
    \texttt{</think>}}
    
    \vspace{0.5em}
    {\color{answercolor}
    \textbf{\texttt{<answer>}
    The skyline in the image is located in Philadelphia, Pennsylvania, USA.
    \texttt{</answer>}}}

\end{tcolorbox}

\definecolor{promptheader}{RGB}{70, 130, 180} 
\definecolor{promptbody}{RGB}{245, 248, 250}   
\definecolor{thinkcolor}{RGB}{105, 105, 105}   
\definecolor{toolcolor}{RGB}{178, 34, 34}      
\definecolor{obsframe}{RGB}{220, 220, 220}     
\definecolor{answercolor}{RGB}{0, 100, 0}
\begin{tcolorbox}[
    enhanced,
    colback=promptbody,
    colframe=promptheader,
    coltitle=white,
    title=\textbf{\faCogs\ Case 2 from LiveVQA},
    fonttitle=\bfseries\large,
    arc=3mm,
    boxrule=0.5mm,
    left=3mm, right=3mm, top=3mm, bottom=3mm,
    standard jigsaw,
    opacityback=1,
    breakable
]

    \textbf{\faQuestionCircle\ Question:} At what precise location did the gathering of participants, as seen in the group photo on the orange carpet, occur during the most recent iteration of this event?
    
    \begin{center}
        \includegraphics[height=3.5cm, keepaspectratio]{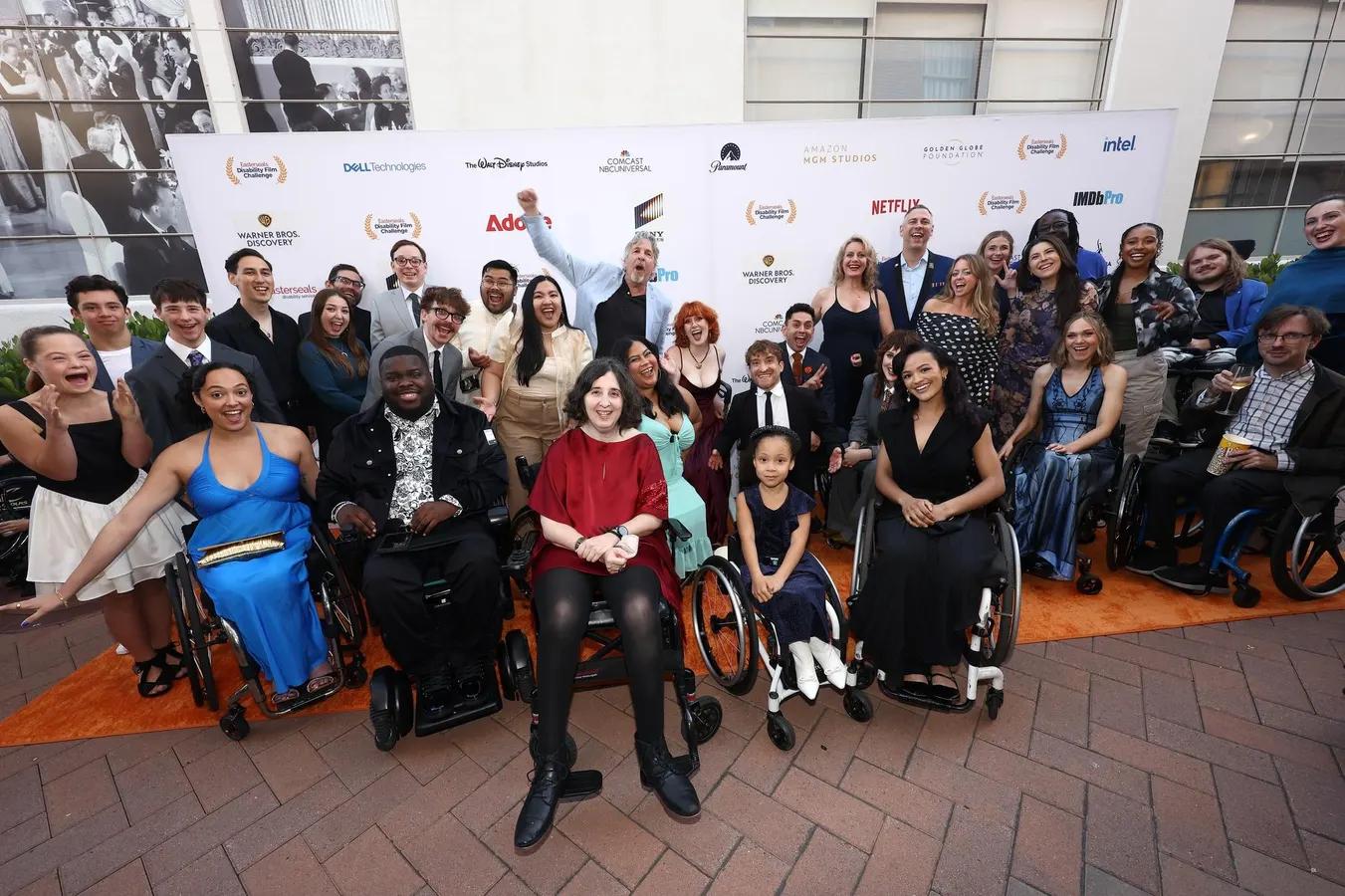}
    \end{center}

    \noindent\rule{\linewidth}{0.4pt} 
    
    \textbf{Step 1}
    
    {\color{thinkcolor}\small \faLightbulb\ \texttt{<think>} The image shows a group of individuals gathered at an event with logos like Adobe, Warner Bros. Discovery... I need to identify details about the event... \texttt{</think>}}
    
    \vspace{0.3em}
    {\color{toolcolor}\small \faSearch\ \textbf{Tool Call:} \texttt{web\_image\_to\_image\_search}}
    
    \begin{tcolorbox}[colback=white, colframe=obsframe, title=\footnotesize \faEye\ Tool Observation (Image Search Results), sharp corners, boxrule=0.3mm, left=1mm, right=1mm, top=1mm, bottom=1mm]
        \footnotesize
        \textbf{Title:} 1. Easterseals Disability Film Challenge Awards... 2. In The News...\\
        \textbf{Retrieved Images:}
\begin{center}
            \begin{minipage}[c]{0.32\linewidth}
                \centering
                \includegraphics[height=2.2cm, max width=\linewidth, keepaspectratio, valign=m]{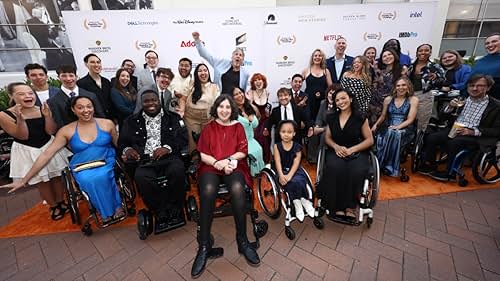}
            \end{minipage}%
            \hfill 
            \begin{minipage}[c]{0.32\linewidth}
                \centering
                \includegraphics[height=2.2cm, max width=\linewidth, keepaspectratio, valign=m]{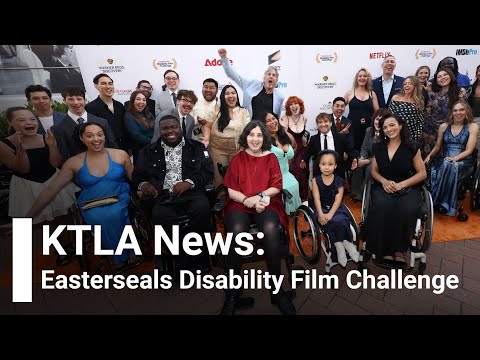}
            \end{minipage}%
            \hfill
            \begin{minipage}[c]{0.32\linewidth}
                \centering
                \includegraphics[height=2.2cm, max width=\linewidth, keepaspectratio, valign=m]{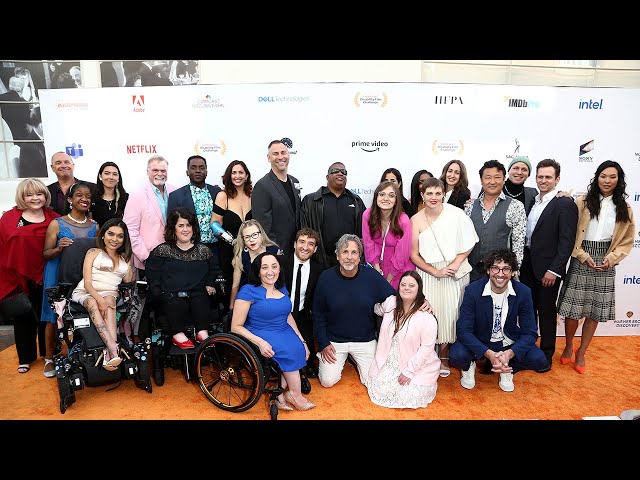}
            \end{minipage}
        \end{center}
    \end{tcolorbox}

    \textbf{Step 2}
    
    {\color{thinkcolor}\small \faLightbulb\ \texttt{<think>} From the image and links... it appears to be the Easterseals Disability Film Challenge... I need to find the exact location... \texttt{</think>}}
    
    \vspace{0.3em}
    {\color{toolcolor}\small \faSearch\ \textbf{Tool Call:} \texttt{search(query=["Easterseals... 2025 location", "Adobe Easterseals... awards ceremony"])}}
    
    \begin{tcolorbox}[colback=white, colframe=obsframe, title=\footnotesize \faList\ Tool Observation (Web Search Results), sharp corners, boxrule=0.3mm, left=1mm, right=1mm, top=1mm, bottom=1mm]
        \footnotesize
        \textbf{Source 3 (Variety):} "...announced the winners of its 2025 edition with an annual orange carpet and awards ceremony on Thursday at \textbf{Sony Pictures Studios}..."\\
        \textbf{Source 4 (Forbes):} "The 2025 Disability Film Challenge awards ceremony take place at a special orange carpet event at \textbf{Sony Pictures Studios located in Culver City, California}."
    \end{tcolorbox}

    \textbf{Step 3}
    
    {\color{thinkcolor}\small \faLightbulb\ \texttt{<think>} The majority of sources indicate that the 2025 Easterseals Disability Film Challenge awards ceremony took place at Sony Pictures Studios... \texttt{</think>}}
    
    \vspace{0.5em}
    {\color{answercolor}
    \textbf{\faCheckCircle\ \texttt{<answer>}
    The gathering of participants in the group photo on the orange carpet occurred at the Sony Pictures Studios in Culver City, California during the 2025 Easterseals Disability Film Challenge awards ceremony.
    \texttt{</answer>}}}

\end{tcolorbox}
\end{document}